\documentclass[10pt,twocolumn,letterpaper]{article}

\usepackage{cvpr}              

\usepackage{graphicx}
\usepackage{amsmath}
\usepackage{amssymb}
\usepackage{booktabs}
\usepackage[dvipsnames]{xcolor}
\usepackage{multirow}
\usepackage{array}
\usepackage{tabularx}
\usepackage{color}
\usepackage{colortbl}
\usepackage{paralist}
\newcolumntype{K}[1]{>{\centering\arraybackslash}p{#1}}
\newcommand{\drop}[1]{\textcolor{gray}{\textsubscript{--#1}}}
\newcommand{\rise}[1]{\textcolor{gray}{\textsubscript{+#1}}}
\newcommand{\Drop}[1]{\textcolor{Red}{\textsubscript{\bf --#1}}}
\newcommand{\Rise}[1]{\textcolor{Green}{\textsubscript{\bf +#1}}}
\newcommand{\band}{\rowcolor{gray!20}}
\newcommand{\bandc}{\cellcolor{gray!20}}

\newcolumntype{?}{!{\vrule width 1.5pt}}

\usepackage{algorithm}
\usepackage{algorithmicx}
\usepackage[noend]{algpseudocode}
\algdef{SE}[DOWHILE]{Do}{doWhile}{\algorithmicdo}[1]{\algorithmicwhile\ #1}

\usepackage[pagebackref,breaklinks,colorlinks]{hyperref}

\usepackage[capitalize]{cleveref}
\crefname{section}{Sec.}{Secs.}
\Crefname{section}{Section}{Sections}
\Crefname{table}{Table}{Tables}
\crefname{table}{Tab.}{Tabs.}

\def\cvprPaperID{4884} 
\def\confName{CVPR}
\def\confYear{2022}

\newcommand{\fswap}[0]{F-SWAP}

\begin{document}



\newcommand{\R}{\mathbb{R}}

\title{SimVQA: Exploring Simulated Environments for Visual Question Answering}

\author{Paola Cascante-Bonilla\textsuperscript{$\dagger$}\thanks{Work partially done while interning at the MIT-IBM Watson AI Lab} \quad Hui Wu\textsuperscript{$\ddagger$} \quad Letao Wang\textsuperscript{$\natural$} \quad Rogerio Feris\textsuperscript{$\ddagger$} \quad Vicente Ordonez\textsuperscript{$\dagger$}\\
\textsuperscript{$\dagger$}Rice University \quad \textsuperscript{$\ddagger$}MIT-IBM Watson AI Lab \quad \textsuperscript{$\natural$}University of Virginia\\
}
\maketitle

\begin{abstract}
Existing work on VQA explores data augmentation to achieve better generalization by perturbing images in the dataset or modifying existing questions and answers. While these methods exhibit good performance, the diversity of the questions and answers are constrained by the available images. In this work we explore using synthetic computer-generated data to fully control the visual and language space, allowing us to provide more diverse scenarios. We quantify the effectiveness of leveraging synthetic data for real-world VQA.  
By exploiting 3D and physics simulation platforms, we provide a pipeline to generate synthetic data to expand and replace type-specific questions and answers without risking exposure of sensitive or personal data that might be present in real images. We offer a comprehensive analysis while expanding existing hyper-realistic datasets to be used for VQA. 
We also propose Feature Swapping (\fswap{}) -- where we randomly switch object-level features during training to make a VQA model more domain invariant. We show that \fswap{} is effective for improving VQA models on real images without compromising on their accuracy to answer existing questions in the dataset.
\end{abstract}
\section{Introduction}
\label{sec:intro}

\begin{figure}[tbh]
\centering
  \includegraphics[width=0.96\linewidth]{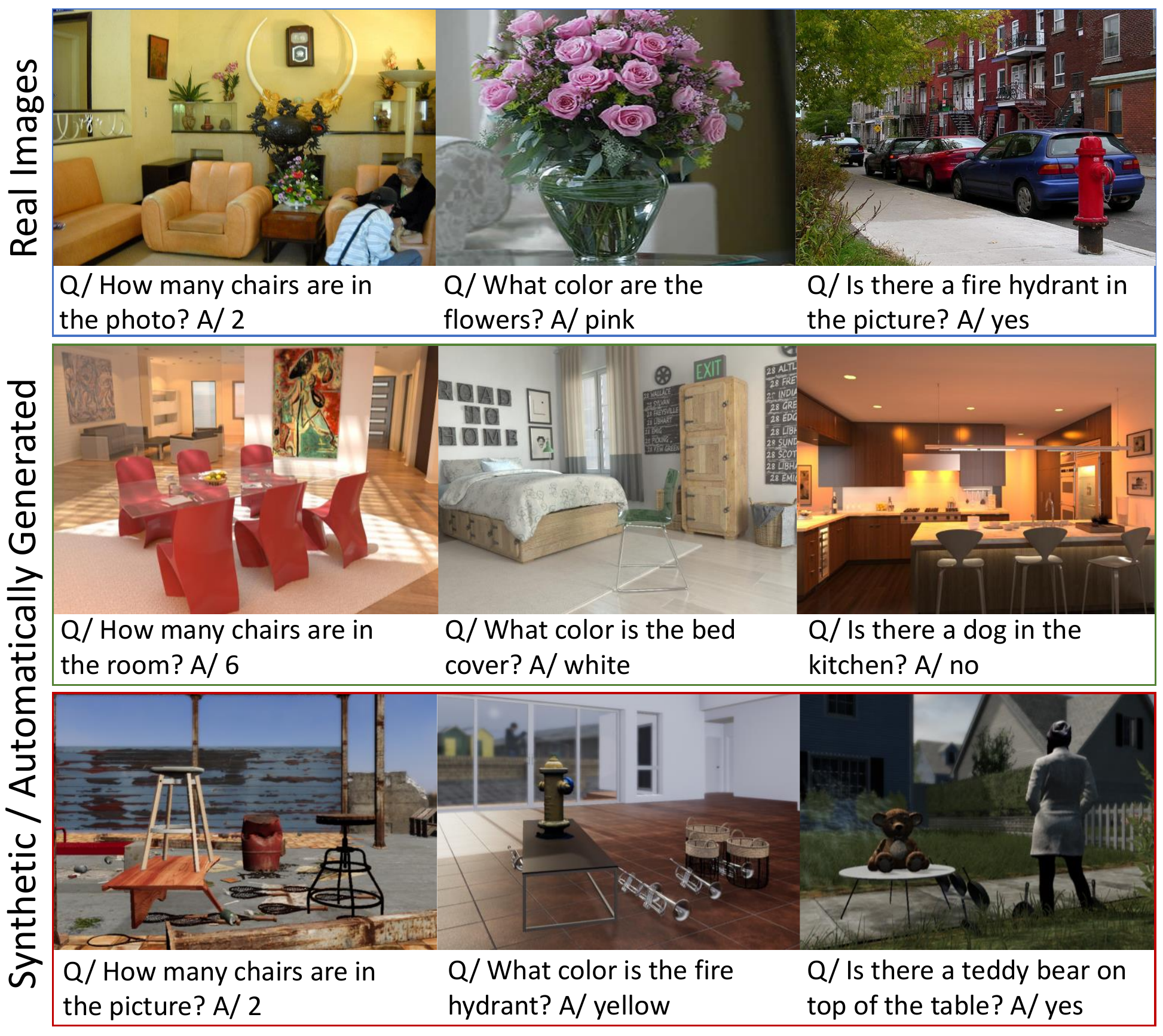}
  \vspace{-0.05in}
\caption{Training samples for VQA from real and synthetic datasets. The first row shows existing examples from the {VQA~2.0} dataset. The second row shows examples from Hypersim~\cite{Roberts2020HypersimAP}, a hyper-realistic synthetic dataset we extend for VQA. The third row shows some examples we generate using ThreeDWorld~\cite{Gan2020ThreeDWorldAP}. We show type-specific questions for each dataset, i.e., counting questions, color related questions, and yes/no questions. }
\vspace{-0.25in}
\label{fig:lead}
\label{fig:num_labeled}
\end{figure}

Data augmentation is an effective way to achieve better generalization on several visual recognition and natural language understanding tasks. Existing work on Visual Question Answering (VQA) has explored augmenting the pool of questions and answers, e.g.~by perturbing or masking some parts of the images~\cite{kafle2017data,tang2020semantic,agarwal2020towards, PatchMix_2021_BMVC}. Moreover, curating large-scale datasets is a laborious task and sourcing images is an expensive process that needs to account for practical issues such as copyright and privacy. Augmenting existing datasets with synthetically generated data offers a path to enhance our existing data-driven models at a lower cost.

Our work focuses on leveraging synthetically generated data through the use of modern 3D generated computer graphics using a couple of novel resources -- Hypersim~\cite{Roberts2020HypersimAP} and ThreeDWorld~\cite{Gan2020ThreeDWorldAP}.
In the past, leveraging synthetic data has proven challenging due the particularly wide domain gap between synthetic images and real images. However, there have been some successes in tasks such as eye gaze estimation~\cite{shrivastava2017learning}, embodied agent navigation~\cite{savva2017minos,deitke2020robothor,savva2019habitat}, and autonomous driving~\cite{prakash2019structured, richter2021enhancing}. 
There have also been some synthetic datasets for visual question answering such as CLEVR~\cite{Johnson2017CLEVRAD}  CLEVRER~\cite{Yi2020CLEVRERCE}, and VQA Abstract~\cite{antol2015vqa}. However these VQA datasets build a closed world that is not designed to generalize to real world images. Remarkably, some recent work has managed to show domain transfer from cartoon images to real images~\cite{Zhang2021DomainrobustVW}, but there is still a limitation on how much could be learned from these existing resources. Our proposed Hypersim-VQA and ThreeDWorld-VQA datasets provide a promising alternative that more realistically captures real world settings and offers a path forward in this direction. Figure~\ref{fig:lead} shows synthetic image samples along with the VQA 2.0 dataset~\cite{goyal2017making}.

Our work also proposes feature swapping (\fswap{}) as a simple yet effective method to augment a currently existing VQA dataset with computer graphics generated examples. Existing methods for domain adaptation rely on the assumption that adaptation can be addressed by making the out-of-domain samples match the distribution of the in-domain samples. However current work often operationalizes this assumption by making the input images themselves look more like the real images e.g.~\cite{tzeng2017adversarial,hoffman2018cycada,rodriguez2019domain}. While there has been success in applying these techniques in domain adaptation for a number scenarios, we claim that perhaps adapting the input space is a harder problem that needs to be solved in order to have effective domain adaptation. Feature Swapping relies instead on swapping random object-level intermediate feature representations. We posit that unless realistic style-transfer is desired from the input domain to the target domain, as long as the two domains are matched at the feature level -- domain adaptation can take place. We explain and compare our \fswap{} approach with other methods such as adversarial domain adaptation and demonstrate superior results.

\vspace{0.04in}
\noindent Our contributions can be summarized as follows:
\begin{compactitem}
\item Dataset generation: We are providing an extension of the Hypersim dataset for VQA, and automatically creating a synthetic VQA dataset using ThreeDWorld.
\vspace{0.04in}

\item Feature swapping (F-SWAP): We propose a surprisingly simple yet effective new technique for incorporating synthetic images in our training while mitigating domain shift. Our method does not rely on GANs or adversarial losses which could be difficult to train.
\vspace{0.04in}

\item Experimental results: We provide an empirical analysis, using well known techniques such as adversarial augmentation, domain independent fusion, and maximum mean discrepancy matching to alleviate the visual domain gap vs our proposed approach -- and analysis on knowledge transfer between skills.
\vspace{0.04in}

\end{compactitem} 

We first introduce related work (Sec.~\ref{sec:related}), then we describe our proposed synthetic dataset generation process (Sec.~\ref{sec:dataset}), then we explain the motivation and details of our feature swapping method (Sec.~\ref{sec:fswap}), then we describe and discuss our experiments(Sec.~\ref{sec:exp}), and finally we conclude the paper (Sec.~\ref{sec:conclusion}). 
Our synthetic datasets and code are available at \href{https://simvqa.github.io}{https://simvqa.github.io.}

\section{Related Work}
\label{sec:related}
Our work is related to both general efforts at improving visually-grounded question-answering models, and efforts targeting data augmentation for VQA and the use of synthetic data for other visual reasoning tasks.

\vspace{0.04in}
\noindent{\bf Visual Question Answering (VQA).} There has been much progress on the task of VQA, where the goal is to answer a question conditioned on both an image and a question text input~\cite{ren2015exploring,antol2015vqa}.
A lot of work in VQA measure progress using the {VQA~2.0} benchmark~\cite{goyal2017making}. Much of the work in this area focuses on exploring
new architectural designs which can effectively model the interaction between the
image and the text modalities, such as bilinear pooling~\cite{fukui2016multimodal}, bottom-up-top-down attention~\cite{anderson2018bottom}, neural module networks~\cite{hu2018explainable, hu2017learning}, and most recently
transformer architectures~\cite{chen2020uniter,li2019unicoder,li2019visualbert}. However, most work assumes that models are trained on real image-question-answer triplets, and that they will be applied to settings with similar data distributions. Our
paper instead investigates a setting where we leverage synthetic training data 
to learn certain skills so that the model generalizes to real images at test time. 

\vspace{0.04in}
\noindent{\bf Data augmentation for VQA.}
Data augmention in VQA has often been studied within the context of model robustness. 
Chen~et~al~\cite{chen2020counterfactual} select visual objects in images and words in questions which
are critical for answer prediction and synthesizes new samples by masking out 
critical visual regions or words. Whitehead~et~al~\cite{whitehead2020learning} leverage existing 
linguistics resources to create word substitution rules for paraphrases, synonyms and antonyms which are then used to generate question perturbations for VQA. Gokhale~et~al~\cite{gokhale2020mutant} explored VQA data synthesis via a combination of semantic manipulation on image content and questions. However, previous work in this direction
generates new samples via perturbations on top of the original real-image VQA dataset, which limits the diversity and the range of variations for generated samples. 
In contrast, our work leverages photo-realistic, multi-physics synthetic environments, 
and is able to generate rich image-question pairs parameterized by scene/room types, camera view, object dimensions, object counts and object orientations. In concurrent work Gupta~et~al~\cite{gupta2022swapmix} propose feature swapping for avoiding contextual bias in visual question answering.

\vspace{0.04in}
\noindent{\bf Synthetic data using simulated environments.} The value of leveraging simulated environments to augment training has been 
explored in various vision tasks, such as object detection, semantic segmentation, and pose estimation~\cite{tremblay2018falling, Salas2020TrainingWS, Zanella2021AutogeneratedWD, Meloni2021SAILenvLI, JohnsonRoberson2017DrivingIT,Le2021EDENMS}. Synthetic environments
have also been applied to vision and language problems, such as embodied agent learning~\cite{Duan2021ASO, Savva2019HabitatAP, Kolve2017AI2THORAI, GarciaGarcia2018TheRA, Szot2021Habitat2T, Ehsani2021ManipulaTHORAF}, using platforms such as the Unreal Engine \cite{MartinezGonzalez2019UnrealROXAE, Qiu2017UnrealCVVW}, and using existing scenes and spaces manually created by specialized designers and content creators~\cite{UE4Arch}. 
Within the task of VQA, to train and diagnose model performance on compositional questions, synthetic datasets such as CLEVR~\cite{Johnson2017CLEVRAD} and CLEVRER~\cite{Yi2020CLEVRERCE} have been proposed. However, models trained on such synthetic datasets typically do not generalize to real images as they were designed under a closed world assumption.
In this paper, we explore approaches which leverage both real-image VQA data for its richness in visual concepts 
and question types, as well as synthetic datasets generated from controllable, configurable 3D environment. We found that using this approach we can generate arbitrarily large-amounts of high-quality data for type-specific 
questions.

\section{Synthetic Dataset Generation}
\label{sec:dataset} 

\begin{figure}[t!]
\centering
  \includegraphics[width=\linewidth]{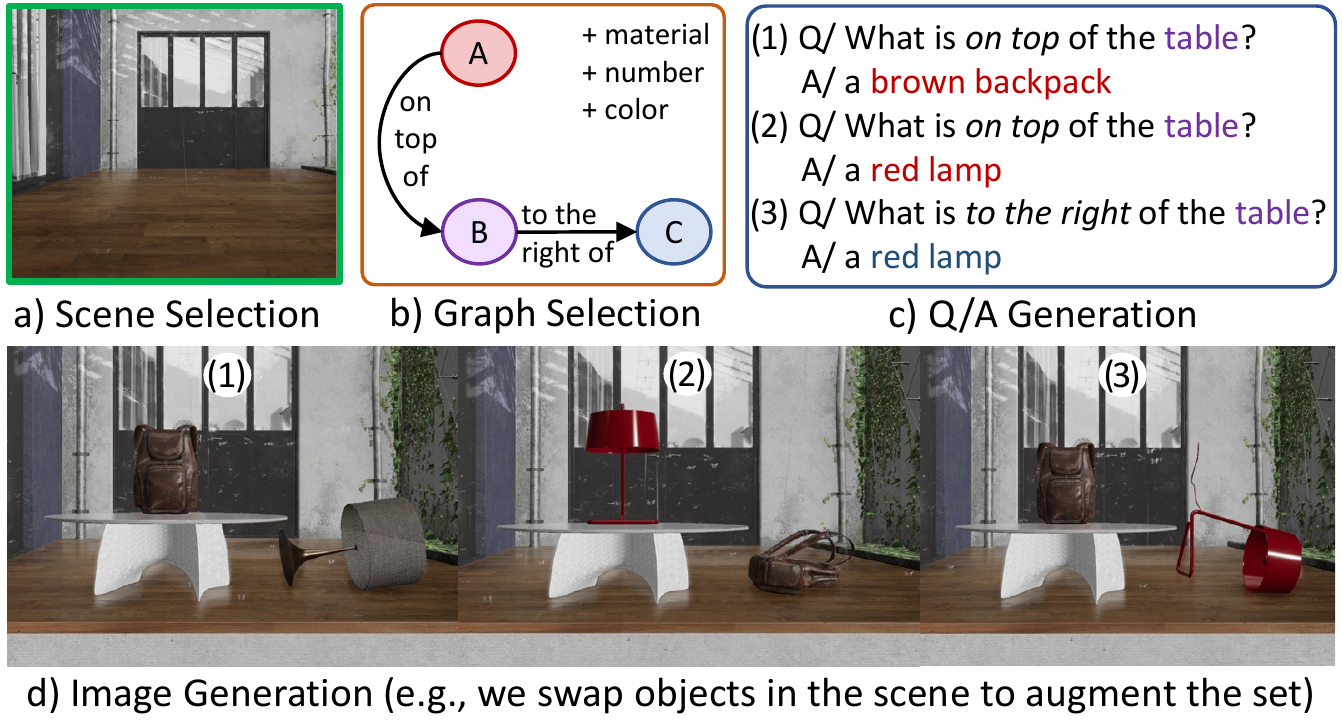}
  \vspace{-0.15in}
\caption{Sample pipeline for generating VQA data using ThreeDWorld. a) Manually select scenes from a set of random camera walks.  
b) Select one of the generated scene graphs containing object information such as positions, number, color, and materials. c) Generate question-answer pairs following a template based on the scene graph. d) Finally, generate images by placing objects and modifying characteristics of the scene based on steps b and c.}
\label{fig:TDW_pipeline}
\vspace{-0.1in}
\end{figure}

First, we describe the generation of a VQA dataset by extending the existing Hypersim dataset~\cite{Roberts2020HypersimAP}~(section~\ref{extending_hypersim_for_vqa}). We name this dataset Hypersim-VQA, or H-VQA, for short. Then we explore the automatic creation of a VQA Dataset using ThreeDWorld~\cite{Gan2020ThreeDWorldAP} (section~\ref{automatic_generation_with_tdw}). We name this dataset ThreeDWorld-VQA, or W-VQA, for short.

\subsection{Extending Hypersim for VQA}
\label{extending_hypersim_for_vqa}

Hypersim~\cite{Roberts2020HypersimAP} is an existing 3D graphics generated dataset with a high image quality and displays a diverse array of scenes and objects. Hypersim metadata includes the complete geometry information per scene, dense per-pixel semantic instance segmentations for every image, and instance-level NYU40 labels annotations. We extend these data by manually annotating objects on all images given their dimensions and positions in the scene. Additionally, we add questions and answers based on the number of appearances of an object in an image and their location with respect to other objects in the same frame. 
Since we have the 3D bounding boxes coordinates for each object in a scene, we can calculate the distance $d$, plunge $p$ and azimuth $a$ for two objects located in positions $(x_1, y_1, z_1)$ and $(x_2, y_2, z_2)$ as 
\begin{small}
\begin{eqnarray}
d &=& \sqrt{(x_2-x_1)^2+(y_2-y_1)^2+(z_2 -z_1)^2}, \\
p &=& \text{deg}\left(\arcsin \left(\frac{z_2-z_1}{d}\right)\right), \\
a &=& \text{deg}\left(\arctan \left(\frac{y_2-y_1}{x_2-x_1}\right)\right),
\end{eqnarray}
\end{small}
where $d$ is the amount of space between two objects, $p$ is the angle of inclination measured from the horizontal axis formed by aligning two objects, and $a$ is the angle between two objects, measured clockwise with respect to the North. Once we have the distance between all objects in a scene, we define clusters of objects and use the azimuth and plunge to define the position of one object with respect to the other objects in the same scene. 
Finally, we generate yes/no questions and answer pairs based in the visibility of an object in a scene frame. We call this new set Hypersim-VQA.

\begin{figure}[t!]
\centering
  \includegraphics[width=\linewidth]{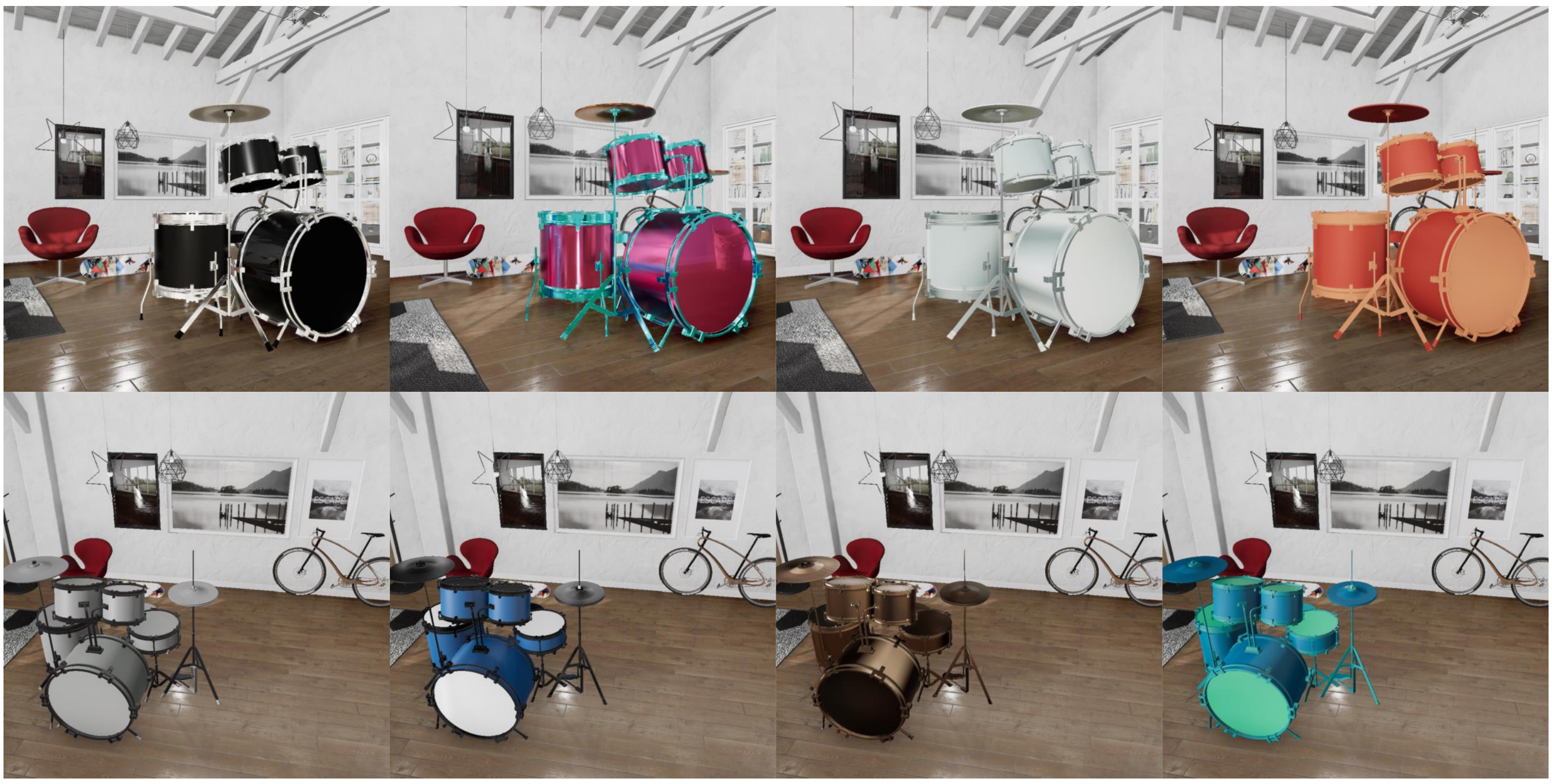}
    \vspace{-0.15in}
\caption{In our VQA dataset generation pipeline, we can automatically manipulate the scene composition, object materials and colors, allowing our grammar to generate more challenging questions and answers. }
  \vspace{-0.1in}
\label{fig:TDW_samples}
\end{figure}

\subsection{Automatic VQA Generation}
\label{automatic_generation_with_tdw}
ThreeDWorld (TDW)~\cite{Gan2020ThreeDWorldAP}, is a platform for interactive multi-modal physical simulation that we use to generate images. We follow the steps shown in Figure~\ref{fig:TDW_pipeline} to generate the image $I$, question $Q$ and answer $A$ triplets for our W-VQA dataset. In this section, we provide a detailed description of the synthetic-generation pipeline.

\vspace{-0.1in}
\paragraph{TDW Model Library}
TDW contains $2,323$ objects, $585$ different materials, and $44$ scenes with $35$ scenes indoors and $9$ outdoors. This variety of assets give us a significant level of freedom to generate diverse and challenging image compositions and question/answer pairs. 
Using the TDW ModelLibrarian\footnote{https://github.com/threedworld-mit/tdw}, we have access to a model asset bounding points which we use to calculate the volume of an object. Using volume information we assign dimension-related categories (e.g., tiny, small, mid-range, large, etc) to each object. We also have access to a set of category labels, that correspond to ImageNet labels, already assigned to each object asset (3D objects). Additionally, we manually annotate the color of each material asset, and added detailed descriptions for each object in the available TDW model set.

\vspace{-0.1in}
\paragraph{Scene graph generation}
We manually designed a set of simple scene graphs~\cite{scenegraphs, johnson2015image} that describe objects, relationships between objects, and the attributes of each object. We assign the position relationship of these objects given the available range of sizes and object model categories, e.g., small and tiny could be on top of tables, chairs, or other large objects, and large objects could be placed in the vicinity of another large object. We showcase an example of this in Figure~\ref{fig:TDW_pipeline}. The number of objects in a scene is randomly selected based on the object size, e.g., we could place dozens of small objects that fit in a camera view, but having dozens of large objects may create occlusions and collisions in the image, and some objects may fall outside the camera view. Finally, the color and material attributes are assigned randomly, but we also generate a large set in which we keep most of the object models with their original attribute values, as we further describe in Section~\ref{dataset_settings}.

\vspace{-0.1in}
\paragraph{Synthetic Image Generation}
We placed multiple cameras with random configurations in the $44$ scenes, by randomly generating $x_c$, $y_c$, and $z_c$ coordinates for camera positions, and $\theta_c$ for directions that the cameras look at. We then manually selected a set of camera configurations that have good views of an empty room, to later place objects in front of them. For example, Figure~\ref{fig:TDW_pipeline} illustrates a scene in which we placed a random table at the left of the image, a random small object (backpack or lamp) on the table, and another random small object on the ground, which follows the scene graph depicted in Step B of the Figure. We also change the material of objects at this stage, and place a random number of objects in the image following the scene graph configuration. An example of how changing materials and colors visually affects a generated image is showcased in Figure~\ref{fig:TDW_samples}. We calculated the positions of these objects relative to the camera using
\vspace{-0.05in}
\begin{eqnarray}
    x &=& x_c + \, r \cos{\theta}, \\
    y &=& y_0 + \, h, \\
    z &=& z_c + \, r \sin{\theta},
\end{eqnarray}
where $x$, $y$ and $z$ are the position coordinates of the placed object, $\theta$ is the direction of the object with respect to the camera, typically within $30$ degrees to $\theta_c$, the direction that the camera looks at, $r$ is the distance between the object and the camera, and $y_0$ is the coordinate of height at the floor level in the scene. 
In addition, since $h$ is an estimated height with respect to the object size, we waited 25 frames for these objects to fall to their natural stationary positions using the TDW physics engine. Finally, TDW allows to capture the RGB images from the camera view along with the id and category per-pixel semantic masks, which we later use to verify the number of objects in the image and avoid object occlusions.

\begin{figure}[t!]
\centering
  \fbox{\includegraphics[width=0.98\linewidth]{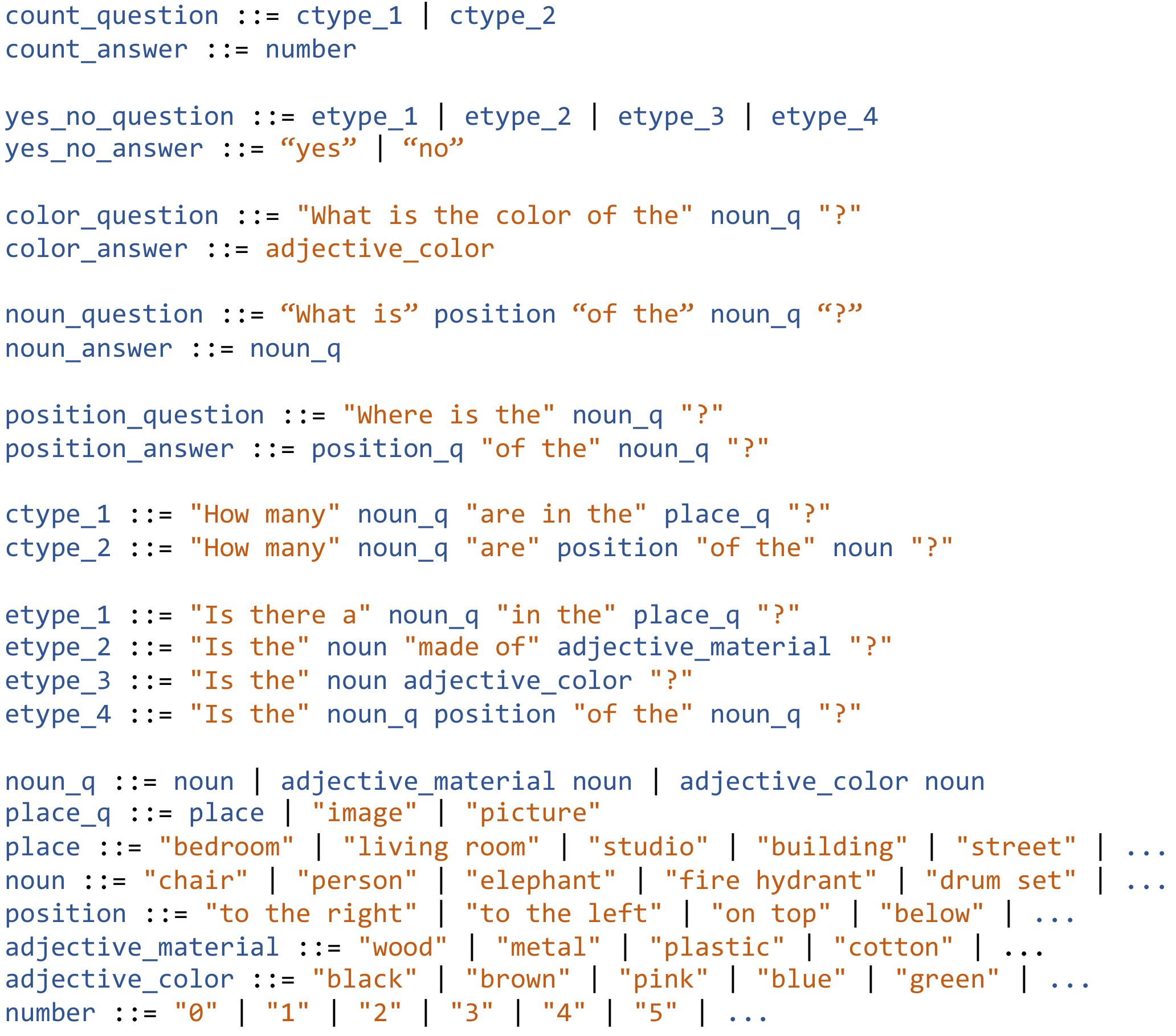}}
  \vspace{-0.1in}
\caption{Template based grammar we use to generate question/answer pairs given our generated image and it's corresponding scene-graph. }
\vspace{-0.1in}
\label{fig:TDW_grammar}
\end{figure}

\vspace{-0.1in}
\paragraph{Question/Answer Generation}
Questions and answers are generated following a template based grammar associated with a predefined scene graph and it's corresponding image. We show in Figure~\ref{fig:TDW_grammar} the template based grammar we use to generate the question/answer pairs. In our setup, a \textit{noun} is directly associated with the model object label category from the TDW asset, \textit{position} is taken from the relationship between objects from the scene graph, and  \textit{number}, \textit{adjective\_color} and \textit{adjective\_material} are taken from the attributes selected when generating the graph and the synthetic image. We refer to the distribution of generated questions and answers with more detail in Section~\ref{dataset_settings}.

\begin{figure*}[t]
\centering
  \includegraphics[width=.99\textwidth]{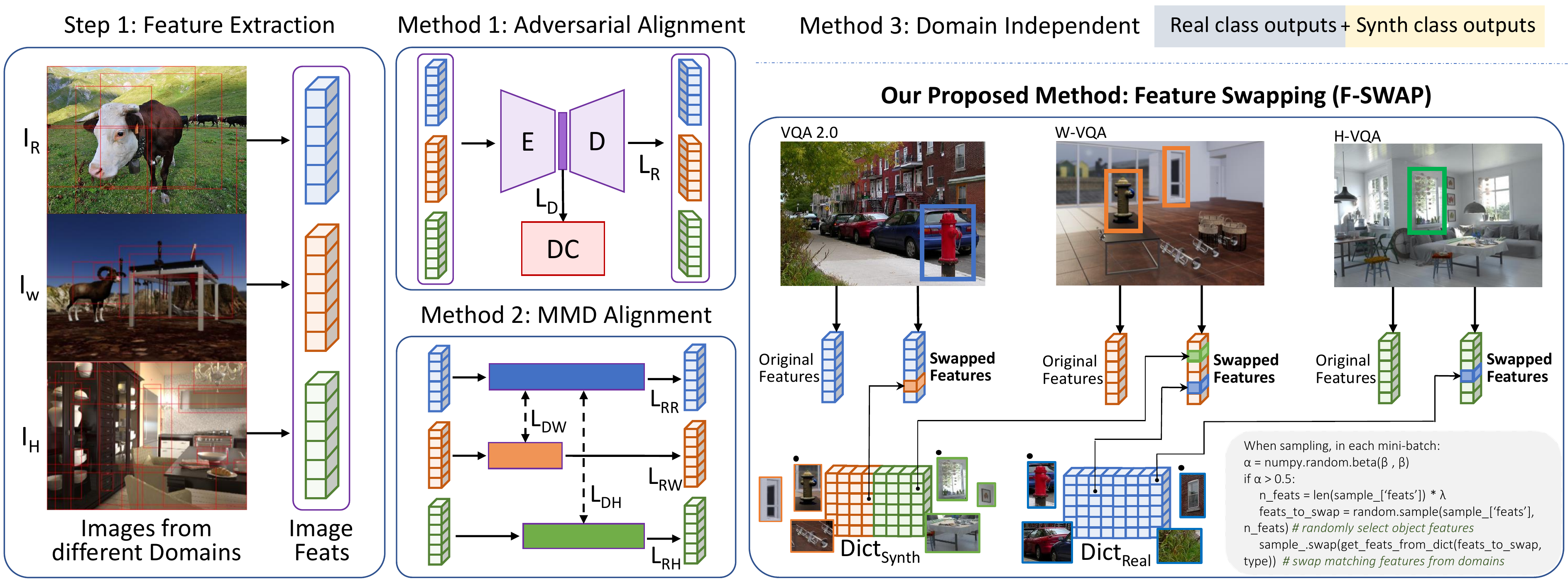}
 \vspace{-0.05in}
\caption{Overview of our training pipeline. First, we extract image the features using Faster-RCNN or CLIP. Then, we use different methods to alleviate the domain gap between real and synthetic images. Methods 1 $\&$ 2 yield a set of aligned features and Method 3 
augment the output space of the VQA model (i.e., answer tokens), separating real class output tokens and synthetic class output tokens. Our proposed method (\fswap{}) swaps object-level features between domains, which are then used to train the VQA model. }
\label{fig:Methods}
\vspace{-0.1in}
\end{figure*}

\section{Feature Swapping}
\label{sec:fswap}

Given a triplet of images $I$, questions $Q$ and answers $A$, we have access to three datasets from different domains, where $(I_R, Q_R, A_R) \in R$ correspond to a Real-VQA dataset consisting of real images (we use {VQA~2.0}\cite{goyal2017making}), $(I_H, Q_H, A_H) \in H$ correspond to the Hypersim-VQA dataset, and $(I_W, Q_W, A_W) \in W$ correspond to the TDW-VQA dataset. We assume that the images and their corresponding questions are inputs to a VQA model, and the objective is to predict as output the corresponding ground-truth answers. Our goal with feature swapping is to train a model that is generic and as domain invariant as possible. The motivation for feature swapping relies in observing that in all three datasets we can find similar types of objects and configurations but the appearance of the objects might differ. Our goal with feature swapping is then to randomly replace during the training the object-level features for some of the objects with the features for an equivalent object from another domain.

Given an image $I$, we use a pre-trained model $G$ to extract the image region features $G_f(I) = \{f_1, f_2, ..., f_n\}$ along with their corresponding pseudo-labels $G_{sl}(I) = \{sl_1, sl_2, ..., sl_n\}$ which are labels predicted by Faster-RCNN corresponding to annotations from Visual Genome~\cite{Krishna2016VisualGC}. Pseudo-labeling has proven effective in semi-supervised learning where only a portion of the training data is annotated\cite{PseudoLabel,arazo2019pseudolabeling,curriculum2021}. Since we have access to all images from the three sets, we create a dictionary $D_{\textit{type}}$ per dataset with $\textit{type} = R \lor H \lor W$, where the $[\textit{key}, \textit{value}]$ of a dictionary $D_\textit{type}$ corresponds to the pseudo-label $(sl_i)_\textit{type}$, and all the region features $[(f_i)_\textit{type}, ..., (f_m)_\textit{type}]$ that the model $G$ assign as $(sl_i)_\textit{type}$ respectively.
Once we retrieve the information for all the dictionaries $D_R, D_H, D_W$, we use them to swap features from one dataset to the other. While training, when sampling datapoints from $R$, we randomly select an Image $I_R$ and get all the region features $G_f(I_R)$ and its corresponding pseudo-labels $G_{sl}(I_R)$. Since we have access to all dictionaries, we lookup for the pseudo-labels that also exist in $D_H \lor D_W$, for simplicity, $D_S = D_H \lor D_W$, thus, after obtaining $G_{sl}(I_R) \in D_S$ we proceed to randomly select a portion $\lambda |G_f(I_R)|$ of the corresponding pseudo-labeled features in $D_S$ and replace them with the matching features in $I_R$. In all of our experiments, $\lambda = 0.2$. Figure~\ref{fig:Methods} shows pseudo-code for this algorithm.

In VQA, the model takes as input pre-computed region features that come from a pretrained object detection model, but the prediction scores are often ignored. VQA models assume that these region features contain enough information for vision and language reasoning. Here we are assuming that the pseudo-labels associated with region features are good predictions; thus, we are augmenting the feature space of the input images from the real domain with features from the synthetic domain by perturbing the input image feature space with features that $G$ scores as similar. In contrast to other methods that rely on adversarial augmentation, our presented approach does not need training or any adaptation. Importantly, since we are performing feature replacements in latent representations we also bypass the need to perform anything resembling style-transfer in the input pixel domain.

\section{Experiments}
\label{sec:exp} 
First, we describe our experimental settings~(Sec.~\ref{dataset_settings}), then we describe our data augmentation experiments~(Sec.~\ref{sec:aug}), then we describe our experiments using various domain alignment techniques including \fswap{}~(Sec.~\ref{sec:domain-alignment}), and finally we show how certain types of question beyond counting can influence the accuracy of counting questions~(Sec.~\ref{sec:questionshelp}).

\subsection{Experimental Settings}
\label{dataset_settings}
\paragraph{Datasets.}
\noindent
\textbf{Real-VQA.} Following Whitehead~et~al~\cite{Whitehead2021SeparatingSA}'s skill-concept separation for compositional analysis, we take the VQA 2.0 dataset~\cite{goyal2017making}, and separate the counting questions for a detailed analysis on how synthetic data may affect a model performance. For training, we create two different splits: R-VQA$_C$ that corresponds to the training set with only counting questions, and R-VQA$_{NC}$ which corresponds to the VQA 2.0 training set without counting questions. R-VQA$_C$ contains $48,431$ datapoints, and R-VQA$_{NC}$ contains $378,018$ datapoints. For testing, we use the standard {VQA~2.0} validation set and report our results on {\it Numeric} questions, where ${\sim}85\%$ of the questions correspond to counting questions, {\it Others}, and {\it Overall} for the general accuracy. 
\textbf{Hypersim-VQA.} The Hypersim dataset~\cite{Roberts2020HypersimAP} comes with annotations corresponding to the NYU40 labels. Additionally, we manually annotate $460$ scenes and add $1,250$ extra labels for objects whose semantic masks has generic annotations, such as \textit{otherstructure}, \textit{otherfurniture} and \textit{otherprop}. We then generate $254,174$ counting questions for $41,551$ images. In our experiments, we use a subset of $20,000$ questions that only containe NYU40 labels (excluding \textit{otherstructure}, \textit{otherfurniture} and \textit{otherprop}) and include $10,000$ randomly selected from the extra annotated labels. We also generate $40,000$ yes/no questions probing whether an object is present in an image
following Section~\ref{extending_hypersim_for_vqa}. \textbf{TDW-VQA.} We generate $33,264$ counting related datapoints and add $30,000$ yes/no questions to the same images. Additionally, we generate $12,000$ extra images and add color and material questions following Section~\ref{automatic_generation_with_tdw}, for a total of $87,264$ automatically generated datapoints using the ThreeDWorld simulation platform.

\vspace{-0.1in}
\paragraph{Base VQA model.} Multi-modal transformer-based architectures 
currently hold the state-of-the-art results on VQA Challenge~\cite{Yu2019DeepMC,chen2020uniter,li2019unicoder}. For our experiments, we select the top-performing model without large-scale pre-training~\cite{Yu2019DeepMC} as our base model. 
Our base code follows the hyper-parameter selection included in their publicly available implementation\footnote{https://github.com/MILVLG/mcan-vqa}. 

\vspace{-0.1in}
\subsubsection{Image Features.}
We experiment with three types of input image features:

\vspace{0.05in}
\noindent
\textbf{Region Features.} We extract intermediate features from a Faster R-CNN model~\cite{Ren2015FasterRT} with ResNet-101 as the backbone, pretrained on the Visual Genome dataset. Following~\cite{Anderson2018BottomUpAT, Yu2019DeepMC}, we obtain a dynamic number of objects $m \in [10, 100]$ by setting a confidence threshold. If the number of objects is lower than $100$, we use zero-padding to fill the final matrix of shape $100 \times 2048$.

\vspace{0.05in}
\noindent
\textbf{Grid Features from CLIP ResNet-50.}  
Following \cite{Jiang2020InDO, Shen2021HowMC}, we use the CLIP model~\cite{Radford2021LearningTV} with the ResNet-50 visual backbone and extract the features from the RoI Pooling layer without any additional fine-tuning. With this approach, we can extract an image representation matrix of size $558 \times 2048$.

\vspace{0.05in}
\noindent
\textbf{Grid Features from CLIP ViT-B.}
We also divide the raw input images into grids of $2 \times 2$, $4 \times 4$ and $8 \times 8$, and use them as inputs for the CLIP ViT-B  model~\cite{Radford2021LearningTV}. We then aggregate the outputs and we use them as an image representation matrix of size $85 \times 552$.

\subsubsection{Textual Features.} Following~\cite{Yu2019DeepMC} we tokenize the input questions into words and transform them into feature vectors using pre-trained 300-dimensional GloVe word embeddings~\cite{Pennington2014GloVeGV}. These embeddings are passed through a one-layer LSTM~\cite{Hochreiter1997LongSM}. We then use all the output features for all corresponding words.

\vspace{-0.1in}
\subsubsection{Domain alignment methods.}
For comparisons to earlier work, we consider the following domain alignment methods that have been proposed in the past either for VQA or for other similar visual recognition problems.

\begin{table}[t]
\newcolumntype{Y}{>{\raggedright\arraybackslash}X}
\newcolumntype{Z}{>{\centering\arraybackslash}X}
\newcolumntype{W}[1]{>{\centering\arraybackslash\hspace{0pt}}p{#1}}

\centering
\footnotesize
\setlength\tabcolsep{0.1pt}
\renewcommand{\arraystretch}{1.0}

\begin{tabularx}{0.98\columnwidth}{l c c c W{0.35in} c W{0.35in} W{0.35in} c Y}

\toprule
{\multirow{3}{*}{\bf Feature backbone}} &~~~&
{\multirow{3}{*}{\bf Feature size}} &~~~& 
\multicolumn{4}{c}{\bf Training data} &~~~&
{\bf R-VQA}\\ 

&&&&
{\bf Real}&&
\multicolumn{2}{c}{\bf Synthetic} && {\bf Accuracy}\\
\cmidrule{5-5}\cmidrule{7-8}\cmidrule{10-10}

&&&&
{\scriptsize \textbf{R}} && 
{\scriptsize \textbf{H}} &
{\scriptsize \textbf{W}} &&
{\it Numeric}\\
\midrule

{\multirow{3}{*}{FRCNN – RN101}}   &&
{\multirow{3}{*}{100$\times$2048}} &&
 \bandc \checkmark &\bandc&\bandc  &\bandc  &\bandc&\bandc 42.73\\
 &&&& \checkmark && \checkmark & && 44.70\Rise{1.97}\\
 &&&& \checkmark &&  & \checkmark && 42.86\rise{0.13}\\
 \cmidrule{1-10}

{\multirow{3}{*}{CLIP - RN50}}  &&
{\multirow{3}{*}{558$\times$2048}} &&
\bandc \checkmark &\bandc&\bandc  &\bandc  &\bandc&\bandc 42.83\\
 &&&& \checkmark && \checkmark &  && 43.61\rise{0.78}\\
 &&&& \checkmark &&  & \checkmark && 42.91\rise{0.08}\\
  \cmidrule{1-10}

{\multirow{3}{*}{CLIP – ViT-B }}  &&
{\multirow{3}{*}{85$\times$512}} &&
\bandc\checkmark &\bandc&\bandc  &\bandc  &\bandc&\bandc 41.93\\
&&&& \checkmark && \checkmark &  && 43.98\Rise{2.05}\\
&&&& \checkmark &&  & \checkmark && 41.35\drop{0.58}\\
 
\bottomrule
\end{tabularx}
\caption{Data augmentation using synthetic data improves Real-VQA performance (R-VQA) on numeric questions, especially when using Hypersim-VQA (H). In all these experiments only counting questions were used for training from both the existing Real-VQA dataset, VQA$_{C}$ (\textbf{R}) and our synthetic dataset variants: Hypersim-VQA (\textbf{H}) and TDW-VQA (\textbf{W}).}
\label{tab:countingonly}
\vspace{-0.05in}
\end{table}

\begin{table*}[t]
\newcolumntype{Y}{>{\raggedright\arraybackslash}X}
\newcolumntype{Z}{>{\centering\arraybackslash}X}

\centering
\footnotesize
\setlength\tabcolsep{1pt}
\renewcommand{\arraystretch}{1.1}

\begin{tabularx}{0.78\textwidth}{l c c c ZcZZ c YYY}

\toprule
{\multirow{3}{*}{\bf Feature backbone}} &~~~&
{\multirow{3}{*}{\bf Feature size}} &~~~& 
\multicolumn{4}{c}{\bf Training data} &~~~&
\multicolumn{3}{c}{\bf\multirow{2}{*}{R-VQA Accuracy}}\\ 

&&&&
{\bf Real}&&
\multicolumn{2}{c}{\bf Synthetic} &&&\\
\cmidrule{5-5}\cmidrule{7-8}\cmidrule{10-12}

&&&&
{\scriptsize R-VQA$_{NC}$} && 
{\scriptsize H-VQA$_C$} &
{\scriptsize W-VQA$_C$} &&
{\it Numeric} &  {\it Others} &  {\it Overall} \\ 
\midrule

{\multirow{4}{*}{FasterRCNN – RN101}}   && 
{\multirow{4}{*}{100$\times$2048}}  
  && \bandc \checkmark &\bandc &\bandc  &\bandc  &\bandc&\bandc 6.08 &\bandc 68.94 &\bandc 60.69\\
 &&&& \checkmark && \checkmark &  && 15.99\Rise{9.91} & 68.97 & 62.02\Rise{1.33}\\
 &&&& \checkmark &&  & \checkmark && 21.18\Rise{15.1} & 68.91 & 62.65\Rise{1.96} \\
 &&&& \checkmark && \checkmark & \checkmark && 24.96\Rise{18.8} & 68.91 & 63.14\Rise{2.45} \\
 \cmidrule{1-12}
 
{\multirow{4}{*}{CLIP - RN50}}  &&
{\multirow{4}{*}{558$\times$2048}}  
 &&\bandc \checkmark &\bandc&\bandc  &\bandc  &\bandc&\bandc 4.55 &\bandc 69.70 &\bandc 61.15\\
 &&&&  \checkmark && \checkmark &  && 10.24\Rise{5.69} & 69.63 & 61.84\rise{0.69}\\
 &&&& \checkmark &&  & \checkmark &&  14.67\Rise{10.12}    &  69.45     &  61.76\rise{0.61}\\
 &&&& \checkmark && \checkmark & \checkmark && 17.81\Rise{13.26} & 69.81 & 62.98\Rise{1.83} \\
  \cmidrule{1-12}
   
{\multirow{4}{*}{CLIP – ViT-B }}  &&
{\multirow{4}{*}{85$\times$512}}  
  &&\bandc \checkmark &\bandc&\bandc  &\bandc  &\bandc&\bandc 5.06 &\bandc 70.12 &\bandc 61.58\\
 &&&& \checkmark && \checkmark &  && 14.06\Rise{9.00} & 70.06 & 62.71\Rise{1.13}\\
 &&&& \checkmark && & \checkmark && 17.25\Rise{12.19} & 70.06 & 63.12\Rise{1.54}\\
 &&&& \checkmark && \checkmark & \checkmark && 20.72\Rise{15.66} & 70.05 & 63.57\Rise{1.99}\\
\bottomrule

\end{tabularx}
\vspace{-0.05in}
\caption{Learning counting skill on real data using data augmentation with our synthetic
datasets. In all these experiments only counting questions were used from our synthetic dataset variants: Hypersim-VQA (H-VQA$_{C}$) and TDW-VQA (W-VQA$_{C}$).}
\vspace{-0.1in}
\label{tab:nocounting}
\end{table*}

\vspace{0.04in}
\noindent\textbf{Adversarial adaptation.} This approach is a modification of the unsupervised domain adaptation of Ganin~et~al~\cite{Ganin2015UnsupervisedDA}. Since our goal is to minimize the feature gap from real $I_R$ and synthetic $(I_W \cup I_H)$ images, instead of using the questions or answers ground-truth to predict the class labels (as the label predictor block), we use an auto-encoder $D(E(\cdot))$ to reconstruct the input features $X$ of the images, and a domain classification model $DC$  that is trained to distinguish the domain of each input. This domain classifier is then connected to the underlying input features $X$ but its gradients are multiplied by a negative constant during training. This gradient reversal layer encourages the features of both domains to remain indistinguishable. This process commonly known as adversarial domain adaptation is optimized in an alternative fashion as follows:
\begin{small}
\begin{equation}
L_D = \sum {{(\log(DC(X)) + \log(1 - \hat{DC}(X)))}},
\end{equation}
\begin{equation}
L_R = \sum (D(E(X))-\hat{D}(\hat{E}(X)))^2,
\end{equation}
\begin{equation}
L_{total} = L_R + \alpha L_D,
\end{equation}
\end{small}
where $L_{total}$ is the loss function to be optimized that encourages a good reconstruction while discouraging the features to encode any domain specific information.

\vspace{0.04in}
\noindent
\textbf{Distribution alignment adaptation.} In this approach, we use $N$ auto-encoder architectures $D(E(\cdot))$ corresponding to the datasets we want to align, (e.g., VQA 2.0 as $R$, TDW as $W$, and Hypersim as $H$) we then compute the Maximum Mean Discrepancy (MMD) loss~\cite{MMDgretton12a} among the intermediate layers of each model. By doing this, the real and synthetic features distributions are encouraged to get closer as similarly used for domain adaptation in~\cite{Tzeng2014DeepDC, Saito2018MaximumCD}. This is performed as follows:
\begin{small}
\begin{equation}
L_{D\diamond} = \text{MMD}(E(X_R), \hat{E}(X_\diamond)),
\end{equation}
\begin{equation}
L_{R\diamond} = \sum (D(E(X_{\diamond}))-\hat{D}(\hat{E}(X_{\diamond})))^2,
\end{equation}
\begin{equation}
L_{total} = L_{R} + \alpha L_{DW} + \beta L_{DH},
\end{equation}
\end{small}
\\
where $\diamond$ can be replaced by $W$ for the TDW features, and $H$ for our extended Hypersim dataset features, and $R$ represents Real-VQA features. Unlike the adversarial domain adaptation approach, here the adversary is not a classifier but a loss that tries to match the distribution of the features across the pair of domains.

\vspace{0.06in}
\noindent
\textbf{Domain independent fusion.}
Inspired by Wang~et~al~\cite{Wang2020TowardsFI}'s work on bias mitigation, we perform domain independent training, where we treat the real and synthetic output space as separate. To do so, we create a new set of classes that contains tokens from the synthetic set only, and extend the real set answer token space with these new tokens, as show in the third method of Figure~\ref{fig:Methods}. This approach can be viewed as two classifiers with a shared backbone that has access to the decision boundary of both the real and synthetic domain.

\begin{table*}[ht]
\newcolumntype{Y}{>{\raggedright\arraybackslash}X}
\newcolumntype{Z}{>{\centering\arraybackslash}X}
\centering
\footnotesize
\setlength\tabcolsep{1pt}
\renewcommand{\arraystretch}{1.2}

\begin{tabularx}{\textwidth}{l l c YYY c YYY c YYY}
\toprule

{\multirow{2}{*}{\bf Data}} & {\multirow{2}{*}{\bf Method}} &~~~&
\multicolumn{3}{@{\hskip 0.22in}c}{\bf +$\textbf{0\%}$ R-VQA$_\textbf{C}$} &~~~& 
\multicolumn{3}{@{\hskip 0.15in}c}{\bf +$\textbf{1\%}$ R-VQA$_\textbf{C}$} &~~~& 
\multicolumn{3}{@{\hskip 0.15in}c}{\bf +$\textbf{10\%}$ R-VQA$_\textbf{C}$} \\

\cmidrule{4-6}\cmidrule{8-10}\cmidrule{12-14}

&&& \textit{Numeric} & \textit{Others} & \textit{Overall} 
&& \textit{Numeric} & \textit{Others} & \textit{Overall} 
&& \textit{Numeric} & \textit{Others} & \textit{Overall} \\

\midrule
\band
H-VQA$_C$ & Simple Augmentation && 
15.99 & 68.97 & 62.02 && 
29.64 & 68.45 & 63.34 && 
35.72 & 68.61 & 64.29\\

H-VQA$_C$ & Adversarial && 
16.07\rise{0.08} & 66.01\Drop{2.96} & 59.46\Drop{2.56} &&
28.31\Drop{1.33} & 66.89\Drop{1.56} & 61.83\Drop{1.51} &&
35.01\drop{0.71} & 66.91\Drop{1.7} & 62.71\Drop{1.58} \\

H-VQA$_C$ & MMD && 24.79\Rise{8.80}& 67.13\Drop{1.84} & 61.58\drop{0.44} &&
31.61\Rise{1.97} & 67.78\drop{0.67} & 63.04\drop{0.30} && 
38.87\Rise{3.15} & 68.36\drop{0.25} & 64.49\rise{0.2} \\

H-VQA$_C$ & Domain Independent && 
22.87\Rise{6.88} & 68.65\drop{0.32} & 62.64\rise{0.62} && 
29.05\drop{0.59} & 68.73\rise{0.28} & 63.52\rise{0.18} && 
37.67\Rise{1.95} & 69.34\rise{0.73} & 65.17\rise{0.88}\\

\midrule

H-VQA$_C$ & Feature Swapping {\footnotesize (\fswap{})} && 
23.38\Rise{7.39} & 69.07\rise{0.10} & 63.07\Rise{1.05} && 
31.64\Rise{2.00} & 69.08\rise{0.63} & 64.15\rise{0.81} && 
39.71\Rise{3.99} & 69.13\rise{0.52} & 65.26\rise{0.97}\\

\midrule
\band
W-VQA$_C$ & Simple Augmentation && 
21.18 & 68.91 & 62.65 && 
31.18 & 68.97 & 64.01 && 
38.47 & 68.86 & 64.87 \\

\midrule

W-VQA$_C$ & Feature Swapping {\footnotesize (\fswap{})} && 
26.84\Rise{5.66} & 68.89\drop{0.02} & 63.67\Rise{1.02} && 
31.21\rise{0.03} & 68.82\drop{0.15} & 63.89\drop{0.12} && 
38.54\rise{0.07} & 68.97\rise{0.11} & 64.97\rise{0.10}\\

\bottomrule
\end{tabularx}
\caption{Counting skill learning under different low-regime settings for Real VQA counting questions (R-VQA$_C$). 
All models share the basic training set: VQA$_{NC}$ (the non-counting subset of VQA v2 training
data).
}
\label{tab:alignment}
\vspace{-0.1in}
\end{table*}

\subsection{Data augmentation experiments}
\label{sec:aug}
First, we evaluate the effect of augmenting Real-VQA data with the proposed synthetic datasets. We are interested to test if the ability of VQA models to answer counting questions on synthetic data could improve the counting performance on real VQA data. We experiment with two different
settings for data augmentation. The first setting tests a scenario where real and synthetic data contain the same question type (in this case, counting questions). 
Table~\ref{tab:countingonly} shows that,
under different feature backbones, the performance of counting questions on 
real data is improved when R-VQA$_{C}$ is augmented with the proposed H-VQA dataset.

The second setting targets a more challenging case, where the real data does not overlap with
the synthetic data in terms of questions types. Specifically, in this setting, for real data, we use R-VQA$_{NC}$, which does not contain counting questions. So the model needs to learn the skill 
for counting questions from the augmented synthetic data alone.
Table~\ref{tab:nocounting} shows that in all settings of feature backbones, and different
combinations of synthetic data augmentations, the model learns to answer counting questions. In this case, augmenting with ThreeDWorld-VQA seems to outperform augmenting with Hypersim-VQA, perhaps
due to a greater extent of controllability of the generated scenes. Lastly, the best results
are obtained by data augmentation using both synthetic datasets.

\subsection{Domain alignment.}
\label{sec:domain-alignment}
As demonstrated in Section~\ref{sec:aug}, counting skills learned from our synthetic datasets can effectively transfer to real VQA data, even when the real training data does not contain counting questions. Here, we explore to what extent skill learning using synthetic data can be helped by explicit alignment of visual features between two domains. The real data used in this experiment includes R-VQA$_{NC}$, as well as R-VQA$_{C}$ under three different regimes ($0\%$, $1\%$, $10\%$).

Table~\ref{tab:alignment} summarizes the experimental results when using different domain alignment approaches. Compared to the baseline method of simple data augmentation, we do not observe an overall improvement with Adversarial Adaptation. Compared to Domain 
Independent, MMD seems to generate more consistent gains on counting questions, across 
various regimes for R-VQA$_C$; however, this gain is also accompanied by decreased 
performance on the split of \textit{Others} and sometimes on the overall evaluation data. 
Finally, the results suggest that Feature Swapping outperforms the baseline and other domain alignment methods, and produces consistent gains on counting
questions as well as the overall accuracy, across different regimes of VQA$_C$.

\subsection{Effect of question distribution.}
\label{sec:questionshelp}
In previous experiments, we focus on augmenting the real dataset with synthetic data
of a specific skill type. In this section, we experiment with increased
diversity of questions on synthetic data, and how it may effect the performance on
different subsets of the real data. As shown in Table~\ref{tab:dist}, on the \textit{Others} category, we observe increased performance when adding more
question types with the TDW-VQA dataset but not with Hypersim-VQA, likely due to the
richer object repository and the more controllable environment of TDW. Interestingly,
for both datasets, adding other question types results in a noticeable gain on the 
counting questions. We hypothesize that these additional questions help with visual
concept learning (on color, object existence, etc), which consequently benefits
counting skill learning since visual concept learning is a basic step to answering counting questions. 

\begin{table}[ht]
\centering
\small
\begin{tabularx}{\columnwidth}{lccc}
\toprule
 \multirow{2}{*}{\textbf{Training data: R-VQA$_{\textbf{NC}}$}} &   \multicolumn{3}{c}{\bf R-VQA Accuracy} \\ 
 
\cmidrule{2-4}
  & {\it Numeric} & {\it Others} & {\it Overall} \\ 
\midrule
H & 15.99 & 68.97 & 62.02 \\
H + Yes/No Questions & 22.11 & 68.38 & 63.17\\
\midrule
W & 21.18 & 68.91 & 62.65 \\
W + Yes/No Questions & 25.43 & 70.10 & 64.24\\
W + Color Questions & 26.98 & 70.24 & 64.56\\
\bottomrule
\end{tabularx}
\caption{Effect of the distribution of synthetic data. We add other type-specific questions to our synthetic data and evaluate its effect on real data.}
\label{tab:dist}
\vspace{-0.1in}
\end{table}

\section{Conclusion}
\label{sec:conclusion}
In this paper we demonstrated the efficacy of VQA datasets generated using 3D computer graphics to incorporate new skills into existing VQA models trained on real data. We particularly showed that we can teach a VQA model how to count objects in the real world by using only synthetic data while not decreasing the model performance on other types of questions. This is challenging since real and synthetic datasets often exhibit a large domain gap. We further proposed \fswap{} as a simple yet effective technique for domain adaptation that is competitive and surpasses previous methods in our experiments.

\section{Broader Impact}
The main ethical aspects of this work have to do with data privacy and mitigating implicit biases in the existing VQA models. In this work, we explored 3D simulation platforms to generate realistic synthetic data as a promising direction to augment or replace existing datasets, effectively avoiding the exposure of potentially sensitive information.
However, our approach is not able to generate a broad diversity of animated objects (e.g., people or animals interacting in the scenes) given that Hypersim only contains indoor scenes, and TDW provides a limited quantity of these type of model assets. While we leverage this information in our generated data, a more complex 3D system would be ideal to craft a more diverse set of animated objects. Therefore, this work is a stepping stone for further explorations to address this issue in the future.
\vspace{0.02in}

\vspace{0.04in}
\textbf{Acknowledgments} 
This work was supported in part by the National Science Foundation under Grants No.~\#2221943 and \#2040961.
\clearpage
{\small
\bibliographystyle{ieee_fullname}
\bibliography{egbib}
}

\clearpage
\appendix
\section{Supplementary Material}
First, we show a list of hyper-parameters and implementation details for all of our adaptation methods in Section~\ref{sec:hyper_param}. Then we show some samples of images and their corresponding per-pixel masks, along with the verification algorithm for counting and occlusions in Section~\ref{sec:id_cat_masks}. Then we show some graph samples from our pool of manually designed scenes for the W-VQA dataset and describe their functionality for our automatic triplet (IQA) generation in Section~\ref{sec:scene_graphs_samples}. Finally in Sections~\ref{sec:wvqa_samples} and \ref{sec:hvqa_samples} we show some samples from W-VQA and H-VQA we randomly select from a diverse set of scenes, with different backgrounds, camera position and illumination. 

\subsection{Hyper-parameter Selection}
\label{sec:hyper_param}

The following are all the hyper-parameter selection for all of our algorithms: 
$lr$ refers to learning rate, $E$ to the number of training epochs, $O$ to the optimizer type, $O_{wd}$ is the optimizer weight decay, $O_\epsilon$ is the term added to the denominator to improve numerical stability, $O_\beta$ are a tuple of coefficients used for computing running averages of gradient and its square. For the Adversarial and MMD methods, the auto-encoder network ($AE$) is trained separately, in a 2 step format following Zhang~et~al.~\cite{Zhang2021DomainrobustVW} Two-stage DA; in both cases the first number in $E$ refers to the training epoch parameter for the $AE$. For Domain Independent, $di_{\text{tokens}}$ is the additional output we use for the synthetic answer tokens.

\begin{table}[h]
\vspace{-0.05in}
\setlength{\tabcolsep}{6.4pt}
\begin{center}
\scalebox{0.75}{
\begin{tabular}{l|l?l|l}
\multicolumn{2}{c}{Adversarial} & \multicolumn{2}{c}{MMD} \\
\hline
$lr = 15e-4$ & $E = 100+13$ & $lr = 1e-3$ & $E = 150+13$ \\
$O_{wd} = 1e-6$ & $O =$ Adam & $O_{wd} = 1e-4$ & $O =$ Adam \\
$O_\epsilon=1e-4$ & $O_\beta=(0.8, 0.8)$ & $O_\epsilon=1e-4$ & $O_\beta=(0.8, 0.8)$ \\
$\alpha = \frac{2} {(1 + exp(-10 * p)) - 1}$ & & $\alpha = 0.4$ & $\beta = 0.6$ \\
\midrule
\multicolumn{2}{c}{Domain Independent} & \multicolumn{2}{c}{F-SWAP} \\
\hline
$lr = 15e-4$ & $E = 13$ & $lr = 15e-4$ & $E = 13$ \\
$O_{wd} = 0.2$ & $O =$ Adam & $O_{wd} = 1e-1$ & $O =$ Adam \\
$O_\epsilon=1e-9$ & $O_\beta=(0.9, 0.9)$ & $O_\epsilon=1e-9$ & $O_\beta=(0.9, 0.98)$ \\
$di_{\text{tokens}} = 100$ &  & $\beta=1.$ & $\lambda=0.2$ \\

\end{tabular}
}
\end{center}
\vspace{-0.15in}
\caption{Hyper-parameter selection details for all methods.} 
\label{tab:class_results1}
\vspace{-0.15in}
\end{table}

\subsection{RGB and Mask Samples}
\label{sec:id_cat_masks}

\begin{figure}[h!]
\centering
  \includegraphics[width=0.96\linewidth]{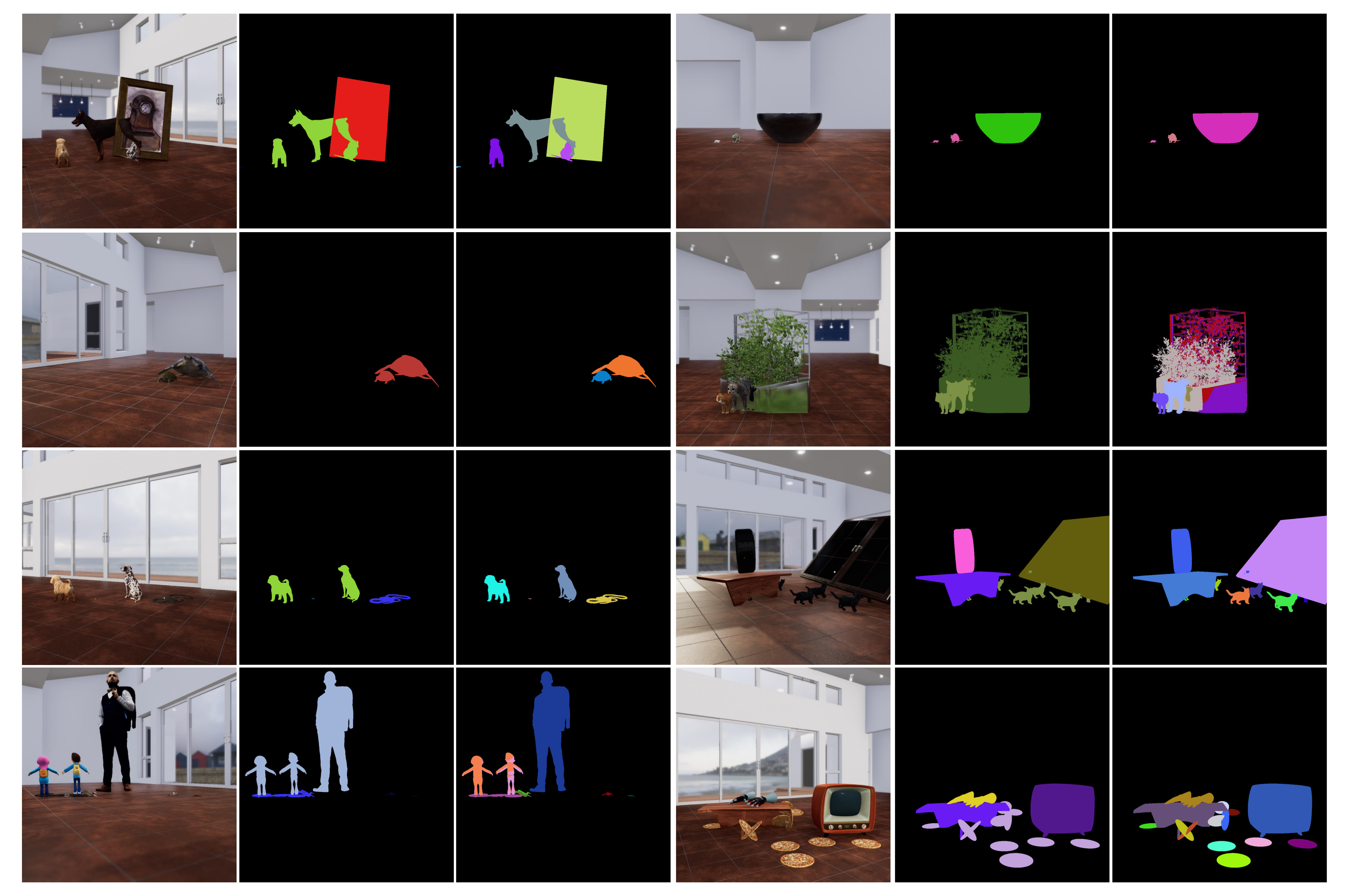}
  \vspace{-0.05in}
\caption{Random samples from the images we generate using TDW along with their category masks (second row) and id masks (third row). }
\vspace{-0.05in}
\label{fig:imgs_and_masks}
\end{figure}

ThreeDWorld (TDW) \cite{Gan2020ThreeDWorldAP}  allows  to capture the RGB images from the camera view along with the  id  and  category  per-pixel  semantic  masks,  which  we later use to verify the number of objects in the image and avoid object occlusions. Figure~\ref{fig:imgs_and_masks} shows some samples we randomly select from our generated W-VQA set. The first column correspond to the RGB image, the second and third columns correspond to the category and id masks respectively.
We verify if an object overlaps to another and assess the object counts by computing the intersection over union. 

\subsection{Scene-Graph Samples}
\label{sec:scene_graphs_samples}
\begin{figure}[h]
\centering
  \includegraphics[width=0.96\linewidth]{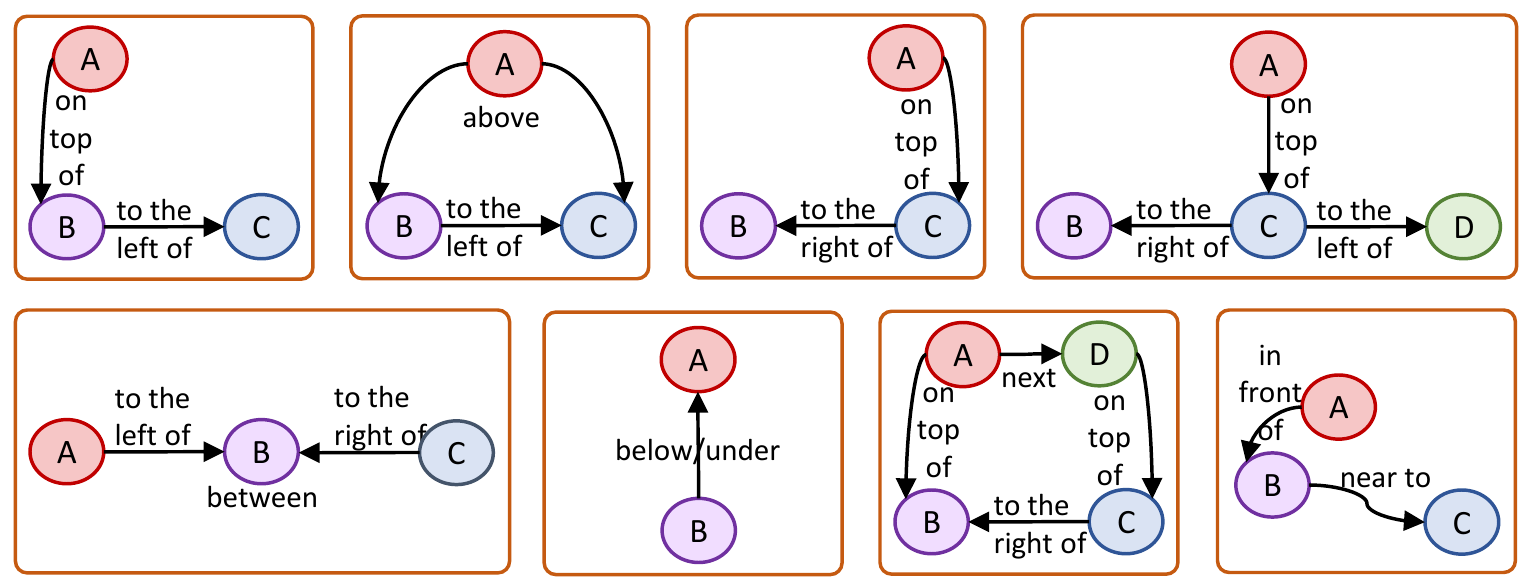}
  \vspace{-0.05in}
\caption{Some of the scene graphs designed for our automated synthetic dataset generation. While generating images, we select one graph and randomly select the number of objects per position $node := [A, B, C, D]$, it's color and materials. Then we use the grammar introduced in Section~\ref{automatic_generation_with_tdw} to generate the questions and corresponding answers.  }
\vspace{-0.05in}
\label{fig:scene_graphs}
\end{figure}

Let $E$ denote the set of scene entities and
consider the set of binary relations $R$. Then a scene graph $SG \in E \times R \times E$ is a collection of ordered triplets $(o, p, o)$ =
object, position, and object. For example, as shown in the first sample in Figure~\ref{fig:scene_graphs}, with $A$=lamp, $B$=table, $C$=backpack, the triplet $(A, position, B)$ indicates that a \textcolor{red}{lamp} is \textcolor{orange}{on top of} the \textcolor{violet}{table}, or the \textcolor{violet}{table} is \textcolor{Mahogany}{under} the \textcolor{red}{lamp}. Similarly, the triplet $(B, position, C)$ indicates that the \textcolor{blue}{backpack} is \textcolor{orange}{to the left of} the \textcolor{violet}{table}, or the \textcolor{violet}{table} is \textcolor{Mahogany}{to the right of} the \textcolor{blue}{backpack}. In this way, from a relationship, there are at least two possible positions, $p \land p^{-1}$, e.g., $p =$ left and $p^{-1} =$ right. When sampling from these graphs, each node in $E$ could also be assigned three different attributes: the number of objects to appear in the same scene $n = randrange(20)$, the color, and material type which are selected from a list of available materials and colors from the set of Records in TDW~\footnote{\url{https://github.com/threedworld-mit/tdw/blob/master/Documentation/misc_frontend/materials_textures_colors.md}}.

\onecolumn

\subsection{W-VQA Generated Samples}
\label{sec:wvqa_samples}
\begin{figure*}[h!]
\centering
  \includegraphics[width=0.75\linewidth]{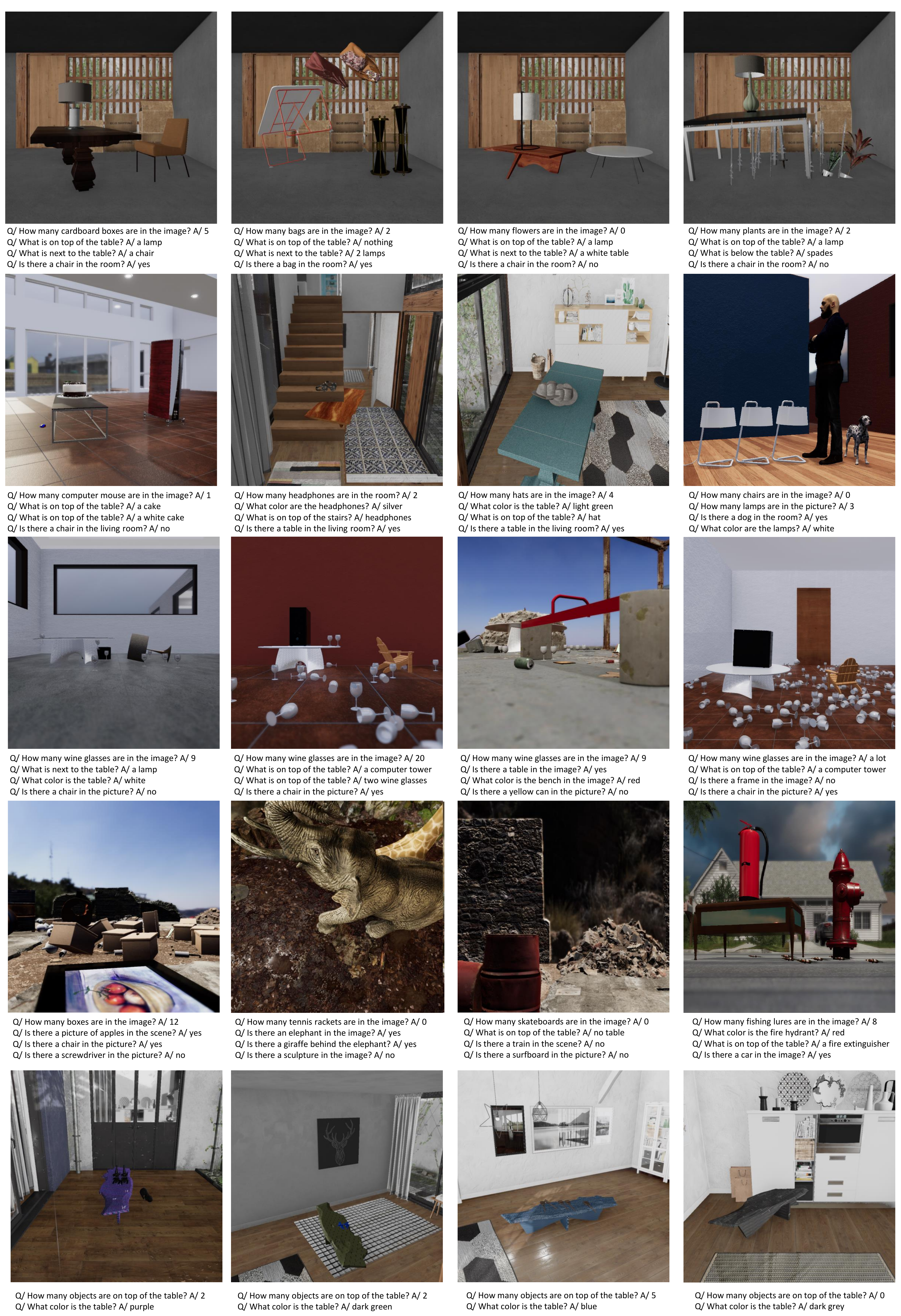}
  \vspace{-0.05in}
\caption{Additional samples of our W-VQA dataset. The first row showcase simple configurations using the same background. The second row shows diverse compositions using indoor scenes. The third row shows compositions of challenging counting questions. The fourth row shows outdoor objects and scenes. Finally, the fifth row shows materials and color related questions using the same object in different camera positions. Best viewed in color.}
\label{fig:fig_W_VQA_extra}
\end{figure*}

We show some random samples we generate for our W-VQA dataset in Figure~\ref{fig:fig_W_VQA_extra}, following Section~\ref{automatic_generation_with_tdw}.

\subsection{H-VQA Generated Samples}
\label{sec:hvqa_samples}
\begin{figure*}[h!]
\centering
\captionsetup{justification=centering,margin=3cm}
  \includegraphics[width=0.77\linewidth]{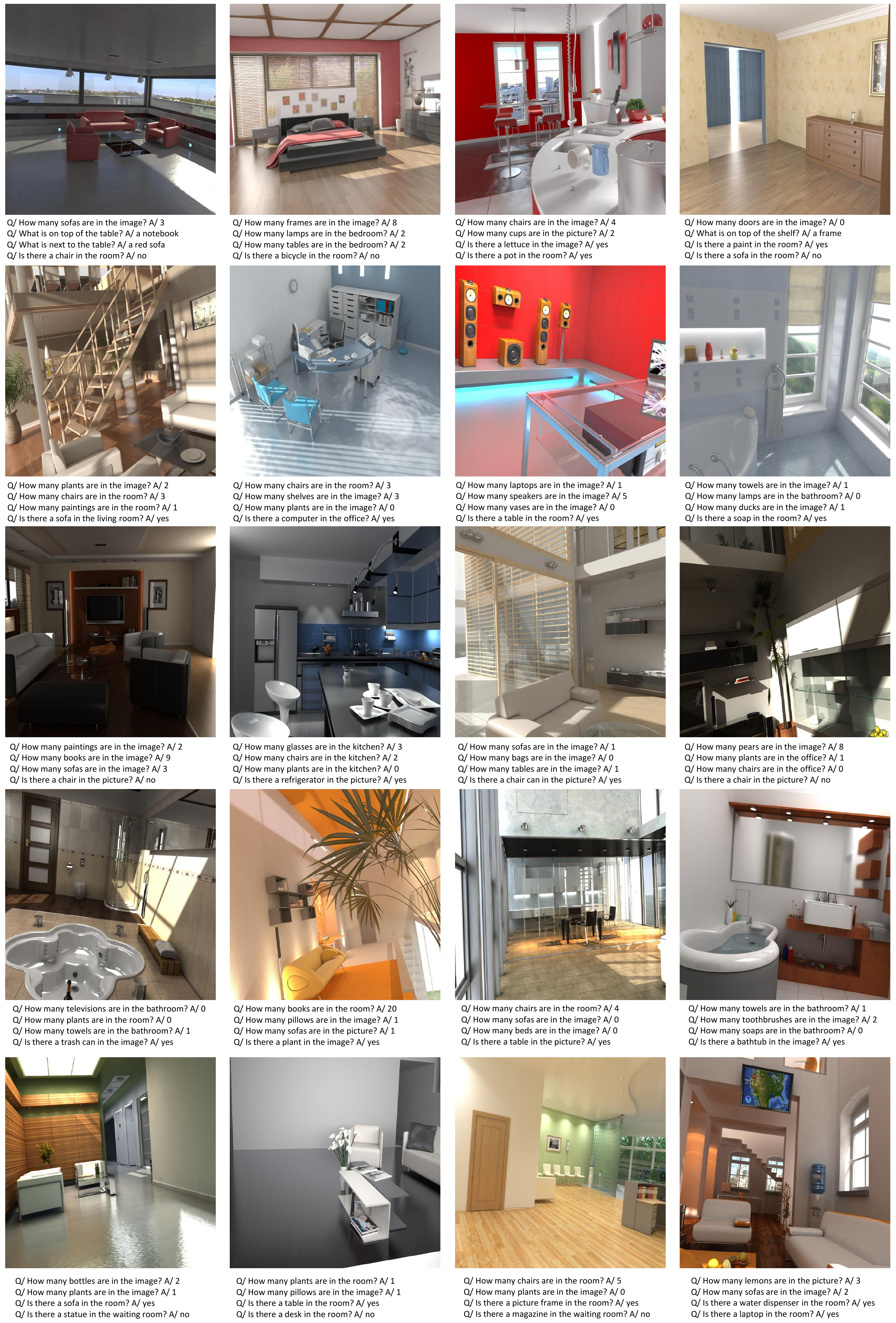}
  \vspace{-0.05in}
\caption{Additional samples of our H-VQA dataset. We generate questions and answers from manual and existing semantic annotations from Hypersim~\cite{Roberts2020HypersimAP}. Best viewed in color. }
\label{fig:fig_H_VQA_extra}
\end{figure*}

We show some random samples we generate for our H-VQA dataset in Figure~\ref{fig:fig_H_VQA_extra}. 

\end{document}